\documentclass[
]{ceurart}

\usepackage{listings}
\usepackage{multirow}
\usepackage{pgfplots}
\usepackage{tikz}
\usepackage{makecell}
\begin{document}

\copyrightyear{2025}
\copyrightclause{Copyright for this paper by its authors.
  Use permitted under Creative Commons License Attribution 4.0
  International (CC BY 4.0).}

\conference{CLEF 2025 Working Notes, 9 -- 12 September 2025, Madrid, Spain}

\title{DS@GT at CheckThat! 2025: Detecting Subjectivity via Transfer-Learning and Corrective Data Augmentation}
\title[mode=sub]{Notebook for the CheckThat! Lab at CLEF 2025}

\author[1]{Maximilian Heil}[
orcid=0009-0002-6459-6459,
email=mheil7@gatech.edu,
]
\cormark[1]

\author[1]{Dionne Bang}[
orcid=0000-0002-9165-8718,
email=dbang6@gatech.edu,
]

\address[1]{Georgia Institute of Technology, North Ave NW, Atlanta, GA 30332}
\cortext[1]{Corresponding author.}


\begin{abstract}
This paper presents our submission to Task 1, Subjectivity Detection, of the CheckThat! Lab at CLEF 2025. We investigate the effectiveness of transfer-learning and stylistic data augmentation to improve classification of subjective and objective sentences in English news text. Our approach contrasts fine-tuning of pre-trained encoders and transfer-learning of fine-tuned transformer on related tasks. We also introduce a controlled augmentation pipeline using GPT-4o to generate paraphrases in predefined subjectivity styles. To ensure label and style consistency, we employ the same model to correct and refine the generated samples. Results show that transfer-learning of specified encoders outperforms fine-tuning general-purpose ones, and that carefully curated augmentation significantly enhances model robustness, especially in detecting subjective content. Our official submission placed us $16^{th}$ of 24 participants. Overall, our findings underscore the value of combining encoder specialization with label-consistent augmentation for improved subjectivity detection. Our code is available at \url{https://github.com/dsgt-arc/checkthat-2025-subject}.
\end{abstract}

\begin{keywords}
  Subjectivity Detection \sep
  Transfer Learning \sep
  Transformer \sep
  Data Generation \sep
  GPT \sep
  Fine Tuning \sep
  CEUR-WS
\end{keywords}

\maketitle

\section{Introduction}
\label{sec:intro}
Given the great risk of misinformation globally\cite{wef2025}, the need for automatic fact-check systems is vital. Similar to machine learning pipelines, an automatic fact-checking system also includes more than just a classifier: retrieval to equip the system with evidence for the fact-check, data preparation to comply with the format requirements of the system, training or fine-tuning to enhance classification performance, and much more. For example, an objective sentence can directly be fact-checked, a subjective sentence needs further data augmentation before it can be passed forward into a fact-checking system. The subjective sentence must be stripped from emotions, opinions or personal interpretations so that the fact-checking system can subsequently focus on the factual verification only. This motivates the CheckThat! Lab of CLEF 2025\cite{Checkthat2025}, where Task 1 focuses on identifying subjective and objective sentences in news paper articles\cite{clef-checkthat:2025:task1}.
Task~1 of CheckThat! has evolved over recent years to address subjectivity detection in multilingual and monolingual contexts. Previous editions in 2023 and 2024 have established strong baselines using transformer-based models and explored both traditional and generative approaches \cite{overview2023, overview2024}. Participating teams have applied lexicon-based classifiers, fine-tuned encoders, and increasingly, synthetic data generation to boost performance under limited data settings. \\
In this paper, we present our contribution to the 2025 English monolingual task. Our approach explores three key research areas focusing on transfer-learning, data-augmentation and the ability of a generative model to correct and refine itself (self-correction).
We evaluate both general pre-trained encoders and compare them with encoders that have already been fine-tuned on related tasks (specified encoders). In addition, we investigate the role of data augmentation through stylistic paraphrasing via a large language model (LLM). Furthermore, we introduce a correction pipeline using the same LLM to align generated paraphrases with their intended labels and stylistic attributes. The impact of each component is assessed through detailed ablation experiments. Overall, our approach to the competition ranks us 16 out of 24. \\
\\
The paper is structured as follows: Section \ref{sec:relateed} highlights related work, Section \ref{sec:method} presents our methodology, Section \ref{sec:data} describes the dataset, Section \ref{sec:res} and Section \ref{sec:disc} shows and discusses the results, Section~\ref{sec:future} highlights future research avenues, and Section \ref{sec:conclusion} concludes.

\section{Related Work}
\label{sec:relateed}
Subjectivity detection has a long history across contexts\cite{10.1007/978-3-540-30586-6_53}, domains\cite{10.1145/3428658.3430978},  and languages\cite{mihalcea-etal-2007-learning}. More specifically, subjectivity detection in news paper articles has a three-year history with CheckThat! at CLEF. Most of the results \cite{overview2023, overview2024} have been driven via the advent of transformer architectures \cite{vaswani2017attention} and the introduction of BERT-like encoder models\cite{devlin-etal-2019-bert} for natural language processing (NLP). NLP systems now easily expand across domains or languages with high accuracy and robustness. In addition, generative models have made substantial contributions due to their outstanding zero-shot and few-shot capability, as well as a significant long context window\cite{gpt2}.

Last year's winner of the monolingual English task, Team DWReCo, employed an LLM to generate subjective training examples with subjectivity styles (e.g. partisan, exaggerated, emotional) given expert knowledge. Synthetic data has been used to balance the data set and enrich the fine-tuning of a \textit{RoBERTa-base}\cite{RobertaBase} model with a classification head.\cite{DWReCo2023} Our data augmentation approach is influenced by Team DWReCo, but we are less interested in data sampling. In contrast, we use stylistic data augmentation for contrastive learning\cite{contrastive}. The 2023 monolingual English task winner, Team HYBRINFOX, builds an ensemble of a fine-tuned \textit{RoBERTa-base} and a \textit{DistilBERT-base-nli-mean-tokens} (sentence transformer\cite{reimers-2019-sentence-bert}) to capture the synthetic as well as semantic meaning of the sentences. This is completed with a self-designed expert system that classifies given NER and lexicon-based methods\cite{HYBRINFOX}. Similarly, we also fine-tune general-purpose encoders on the data set and, in addition, investigate the capability of transfer learning in subjectivity detection.

\section{Methodology}
\label{sec:method}

In the course of the competition, we explored the potential of specialized encoders and data augmentation. As shown in Table \ref{tab:methods}, we contrasted the fine-tuning of general-purpose encoders with transfer-learning of specialized encoders, fine-tuning both on the original data set. In a second step, we explored data augmentation and investigated its added benefit. Finally, we also added a self-corrective data alignment procedure to the data augmentation to ensure that generated paraphrases match their intended labels and styles, using GPT-4o to identify and rewrite those it considers inconsistent.

\begin{table}[h]
\caption{Overview of Explored Methods}
\label{tab:methods}
\begin{tabular}{lll}
\multicolumn{1}{l}{\textbf{Domain}} & \textbf{Model} & \textbf{Dataset} \\ \hline
\multirow{3}{*}{General} & RoBERTa-base & Original Train \\
 & MiniLM-L12-v2 & Original Train \\
 & ModernBERT & Original Train \\
\multirow{3}{*}{Transfer} & Sentiment-Analysis-BERT & Original Train \\
 & Emotion-English-DistilroBERTa-base & Original Train \\
 & Emotion-English-RoBERTAa-base & Original Train \\ \hline
\multirow{2}{*}{Transfer} & Sentiment-Analysis-BERT & Augmented Train \\
 & Emotion-English-DistilRoBERTa-base & Augmented Train \\ \hline
 \multirow{2}{*}{Transfer} & Sentiment-Analysis-BERT & Self-Corrected Augmented Train \\
 & Emotion-English-DistilRoBERTa-base & Self-Corrected Augmented Train \\
\end{tabular}
\end{table}

First, our approach contrasted general-purpose encoders and specialized encoders by evaluating their respective capabilities to distinguish objective from subjective newspaper sentences. Initially, we assessed the performance of general-purpose encoders including \textit{RoBERTa-base}, \textit{MiniLM-L12-v2}, and \textit{ModernBERT}\cite{modernbert}). This evaluation focused on comparing their capabilities in token-level and sentence-level semantic relationships, as well as contextual understanding.
We then employed the transfer-learning capabilities of encoders \textit{Sentiment-Analysis-BERT}\cite{sentimentanalysis}, \textit{Emotion-English-DistilroBERTa-base}, and \textit{Emotion-English-RoBERTa-large} \cite{emotionenglish}, which were already fine-tuned on related tasks with domain-specific datasets for sentiment analysis and emotion recognition. These models are better equipped to detect emotional tone and subjective language, improving their ability to distinguish between subjective and objective statements.\\ 
\\
\textbf{Hypothesis H1}: Transfer-learning with specialized encoders will result in greater sensitivity in distinguishing between subjective and objective language expressions.
\\ 

Second, we investigated the added benefit of data augmentation. Given the small size of the original dataset, we pursued a strategy to synthetically augment and expand the training data using GPT-4o\cite{OpenAI_GPT4_2023}. Inspired by ClaimDecomp\cite{claimdecomp2022}, which decomposed complex political claims into literal and implied sub-questions to improve fact verification, we hypothesized that generating stylistic paraphrases of labeled sentences could improve classification performance by increasing training diversity. 

ClaimDecomp evaluated sub-question generation using encoder-based language models but found that while literal questions were tractable, implied ones remained challenging due to the models’ limited reasoning and contextual capabilities. They also noted that larger language models like GPT could be more suitable for handling these implicit inferences. In parallel, the CLEF CheckThat! 2024 overview\cite{overview2024} found that transformer-based classifiers augmented with domain-informed synthetic data outperformed baselines and that models consistently struggled more with identifying subjective sentences across languages.

Drawing on these findings, we adopted a generation approach similar in spirit to CLaC-2\cite{clac2024}, who used zero-shot GPT-3 to generate two paraphrases per sentence and labeled them via majority vote. However, rather than perform on-the-fly classification, we used GPT-4o in a few-shot setup to generate paraphrases with controlled styles, both subjective and objective, based on the original sentence content. Unlike DWReCo\cite{DWReCo2023}, who used a zero-shot prompt method to generate new subjective examples based on a subjectivity style checklist, we generated both subjective and objective sentences to augment each data point. DWReCo also found that augmentation with paraphrased sentences produced lower-diversity samples and required post-hoc filtering. We did not apply such filtering in our submitted results due to time constraints.

Our generation approach was guided by the hypothesis that providing both subjective and objective perspectives per original content could improve a model's ability to distinguish stylistic cues in spirit of contrastive learning. For each subjective sentence, we generated two or six objective paraphrases. Conversely, for each objective sentence, we generated two or six subjective variants, explicitly styled as \textit{propaganda}, \textit{exaggerated}, \textit{emotional}, \textit{derogatory}, \textit{partisan}, or \textit{prejudiced} (categories aligned with DWReCo’s most effective style prompts).

\begin{itemize}
    \item Original (SUBJ): ``Gone are the days when they led the world in recession-busting.''  
    \item Generated (OBJ): ``The era in which they were at the forefront of overcoming economic downturns has ended.''
    \\
    \item Original (OBJ): ``The trend is expected to reverse as soon as next month.''  
    \item Generated (SUBJ): ``A promising turnaround is on the horizon, with expectations for change as early as next month.''
\end{itemize}

This resulted in two augmented datasets:
\begin{itemize}
    \item \textbf{Balanced-2}: Each sentence augmented with 2 paraphrases in the opposite style
    \item \textbf{Balanced-6}: Each sentence augmented with 6 paraphrases covering a wider stylistic range
\end{itemize}

To assess the impact of this augmentation, we selected two encoder models for fine-tuning: \textit{Sentiment-Analysis-BERT} and \textit{Emotion-English-DistilRoBERTa-base}. These were chosen based on their stronger baseline performance compared to general-purpose encoders, which underperformed consistently in earlier trials and was excluded from further experiments. \\
\\
\textbf{Hypothesis H2}: Fine-tuning with augmented train data will improve the classification performance over the original data set.
\\ 

After submission, in attempt to further improve the quality of our augmented datasets, we developed a second-stage validation and correction pipeline using GPT-4o. Our goal was to ensure that each generated sentence not only aligned with its assigned label (\textbf{SUBJ} or \textbf{OBJ}), but also reflected the appropriate stylistic intent. After visual inspection of the generated samples we recognized that some synthetic data samples were misleading and could degrade the performance when used in fine-tuning the classifier. 
Therefore, we implemented an automated revision program that iterates over each sentence in the augmented training set. For each sample, GPT-4o was prompted with the original sentence, its intended label, and associated style (e.g., \textit{partisan}, \textit{emotional}). The model was instructed to do nothing if the sentence already matched the label and style. Otherwise, it rewrote the sentence to meet the intended classification and style requirements, preserving the subject matter and keeping outputs under 25 words. This method ensured consistency in label-style alignment and improved stylistic clarity without introducing content drift.
The system was built in Python using LangChain's \texttt{ChatOpenAI}\cite{langchain} wrapper and Polars\cite{polars} for input/output. The prompt template emphasized minimal intervention:
\begin{quote}
\textbf{Correction Data Augmentation Prompt:}
\ttfamily
You are an expert in rewriting sentences to match specific subjectivity and style requirements. \\
 \\
Instructions:\\
- You will be given a sentence and its intended label ("SUBJ" or "OBJ") and style. \\
- If the sentence already matches the label and style, return it unchanged. \\
- If it does NOT match, rewrite the sentence so it reflects the correct label and style. \\
- Always preserve the subject matter. \\
- Only apply style if the label is SUBJ. For OBJ, remove all subjective language and opinion. \\
- Keep the rewritten sentence **under 25 words**. \\
 \\
Now perform the task: \\
Label: \{label\} \\
Style: \{style\} \\
Sentence: "\{sentence\}" \\
 \\
Response: \\
\end{quote}
The model operated asynchronously with a concurrency cap to process thousands of samples efficiently, with built-in error handling and whitespace normalization. While the correction step added complexity, the pipeline remained efficient and reproducible. It was run locally using the OpenAI API through LangChain, with GPT-4o rewriting only when it detected a mismatch between a sentence’s content and its assigned label or style. The process completed in a few minutes in practice and outputs were saved as .tsv files with no manual edits. The full pipeline also ran on Georgia Tech’s PACE cluster, ensuring consistent results under controlled API and compute conditions.\\

This process yielded two corrected datasets:
\begin{itemize}
    \item \textbf{Corrected Balanced-2}: Based on the original Balanced-2 augmentation, where each sentence had two style-flipped paraphrases.
    \item \textbf{Corrected Balanced-6}: Based on Balanced-6, with six paraphrases per sentence spanning multiple style prompts.
\end{itemize}

\begin{table}[h]
\caption{Examples of generated sentences deemed mislabeled by GPT-4o}
\label{tab:selfcorrected-examples}
\begin{tabular}{p{0.48\textwidth} p{0.48\textwidth}}
\textbf{Initial Generation (Label)} & \textbf{Corrected Version (Rationale)} \\
\hline
The plight of Serbia’s LGBTQ+ community remains largely unaddressed, leaving them in a void of neglect. (SUBJ, exaggerated) & Serbia’s LGBTQ+ community is shockingly ignored, casting them into an abyss of utter neglect! (increased rhetorical intensity for the exaggerated category) \\
\hline
A promising turnaround is on the horizon, with expectations for change as early as next month. (SUBJ, propaganda) & A glorious transformation awaits us, with change destined to arrive as soon as next month! (intensified tone to better fit propaganda category) \\
\hline
He expressed that a new variant emerging this fall would not come as a shock to him. (SUBJ, propaganda) & The emergence of a new variant this fall is inevitable and will not surprise the vigilant. (rewritten to sound more declarative and assertive to better fit propaganda category) \\
\end{tabular}
\end{table}

We then fine-tuned \textit{Sentiment-Analysis-BERT} and \textit{Emotion-English-DistilRoBERTa-base} on these cleaned and enhanced datasets. As can be seen in Table~\ref{tab:results-aug}, this refinement led to improved macro F1 scores and more stable class-wise performance, particularly in subjective detection. \\
\\
\textbf{Hypothesis H3}: Self-corrected data-augmentation increases the quality of synthetic data and therefore the fine-tuned classifier performance.
\\ 

All results of these described methods can be found in Section \ref{sec:res}.

\subsection{Evaluation}
Our models were fine-tuned and evaluated on the macro-averaged F1-measure (macro F1):
\begin{equation}
   \text{Macro-F1} = \frac{1}{N} \sum_{i=1}^{N} F1_i
\end{equation}
where $F1_i$ is the the class-wise F1 score:
\begin{equation}
    F1_i = 2 \times \frac{P_i \times R_i}{P_i + R_i}
\end{equation}
where $R_i$ is the recall of class $i$ and $P_i$ is the precision of class $i$.

Training was performed on a Tesla V100-16GB on the Phoenix cluster of Georgia Tech's Partnership for an Advanced Computing Environment\cite{Pace} or locally on an Apple M3 Pro GPU-36GB and Metal Performance Shaders.

\section{Data}
\label{sec:data}

\begin{figure}[h]
\label{fig:stats}
\centering
\begin{tikzpicture}
\begin{axis}[
    ybar,
    bar width=0.5cm,
    width=12cm,
    height=7cm,
    ylabel={Number of Sentences},
    xlabel={Dataset Split},
    symbolic x coords={Train, Dev, Dev-test},
    xtick=data,
    ymin=0,
    ymax=600,
    enlarge x limits=0.2,
    legend style={at={(0.5,-0.2)}, anchor=north, legend columns=-1, font=\small},
    nodes near coords,
    nodes near coords align={vertical},
    every node near coord/.append style={font=\small},
    axis line style={draw=none},
    tick style={draw=none},
    major grid style={draw=gray!20},
    grid=major,
    xtick style={draw=none},
    tick label style={font=\small},
    label style={font=\small},
    legend image code/.code={
        \draw[#1,draw=none] (0cm,-0.1cm) rectangle (0.3cm,0.15cm);
    },
]
\addplot [ybar, fill=blue!70] coordinates {(Train,532) (Dev,222) (Dev-test,362)}; \addlegendentry{Objective (OBJ)}
\addplot [ybar, fill=red!70] coordinates {(Train,298) (Dev,240) (Dev-test,122)}; \addlegendentry{Subjective (SUBJ)}
\end{axis}
\end{tikzpicture}
\caption{Class distribution across the English dataset splits}
\label{fig:class-distribution}
\end{figure}
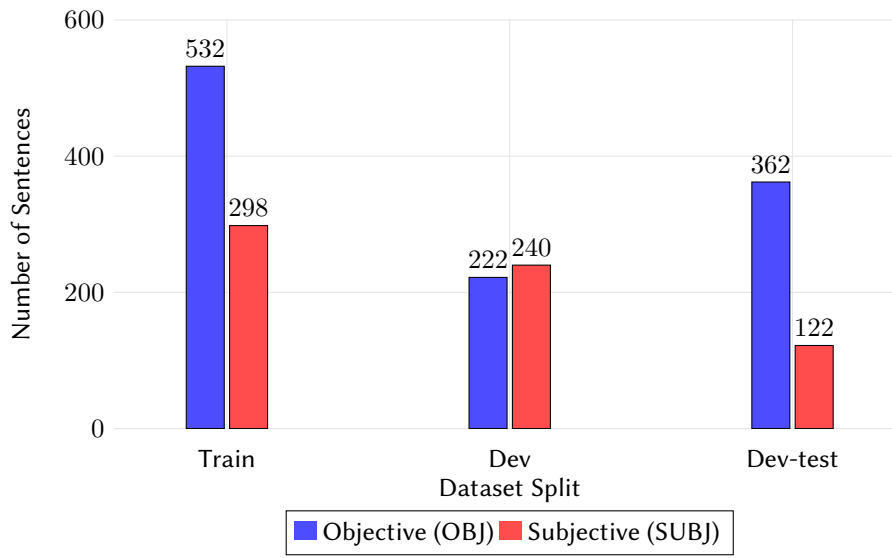

We used the English dataset provided for Task 1 of CheckThat! 2025, which consisted of labeled sentences from news articles. Each sentence was annotated as either \textbf{objective (OBJ)} or \textbf{subjective (SUBJ)}. The dataset was divided into training, development (dev), and test (test) sets. The distribution of class labels across these splits is shown in Figure~\ref{fig:stats}.

The train dataset had substantially more objective sentences (523) instead of subjective sentences (298), but this difference was insufficient to be determined as a significant class imbalance. This did not align with the dev set, which we used as our validation set. Here, the number of objective (222) and subjective (240) examples was more balanced.  
Below are representative examples from each class:

\begin{itemize}
    \item \textbf{Subjective:} ``Gone are the days when they led the world in recession-busting.''
    \item \textbf{Objective:} ``The trend is expected to reverse as soon as next month.''
\end{itemize}

The subjective sentence showed a personal interpretation or opinion, reflecting the speaker's sentiment toward a past event. In contrast, the objective sentence presented factual information or predictions without personal bias. This distinction illustrates that subjective statements often involve emotional or evaluative language, while objective statements rely on logical reasoning.

\section{Results}
\label{sec:res}

Table \ref{tab:results-org} highlights our main results for the general-purpose encoders and domain-specific encoders that we used. \textit{RoBERTa-base} achieved 0.70 macro-F1 on train, 0.75 macro-F1 on validation, and 0.65 macro-F1 on test after fine-tuning for 3 epochs with a learning rate $\alpha$ of 1e-4 on the original train data set. During validation, the \textit{RoBERTa-base} model demonstrates a balanced ability to identify objective and subjective sentences. However, this dropped during test, where the subjective F1 (0.58) was significantly lower than objective F1 (0.72). 

\begin{table}[h]
\caption{Results After Fine-Tuning with Original Train Data}
\label{tab:results-org}
\begin{tabular}{l|c|ccc|ccc}
\multicolumn{1}{c}{\multirow{2}{*}{\textbf{Model}}} & \textbf{Train} & \multicolumn{3}{c}{\textbf{Validation F1}} & \multicolumn{3}{c}{\textbf{Test F1}} \\
\multicolumn{1}{c}{} & \textbf{Macro F1} & \textbf{Obj.} & \textbf{Subj.} & \textbf{Macro} & \textbf{Obj.} & \textbf{Subj.} & \textbf{Macro} \\ \toprule
ModernBERT-large & 0.41 & 0.64 & 0.00 & 0.32 & 0.84 & 0.00 & 0.42 \\
RoBERTa-base & 0.79 & 0.77 & 0.72 & 0.75 & 0.72 & 0.58 & 0.65 \\
MiniLM-L6-v2 & 0.81 & 0.58 & 0.75 & 0.69 & 0.76 & 0.51 & 0.64 \\ \hline
Emotion-english-RoBERTa-large & 0.97 & 0.73 & 0.77 & 0.77 & 0.76 & 0.59 & 0.67 \\
Emotion-english-DistilRoBERTa-base* & 0.90 & 0.78 & 0.75 & 0.77 & 0.80 & 0.57 & 0.68 \\
Sentiment-Analysis-BERT & 0.87 & 0.71 & 0.53 & 0.64 & 0.77 & 0.58 & 0.67
\end{tabular}
\end{table}

\textit{MiniLM-L6-v2} showed weaker generalization across splits, achieving 0.81 macro-F1 on train, 0.69 on validation, and 0.64 on test after fine-tuning for 3 epochs with a learning rate of $\alpha$ 1e-4. It had difficulties with the recognition of objective sentences during validation (only 0.58 F1), though test performance saw a drop in subjective F1 (0.51) compared to objective F1 (0.76), indicating a slight bias toward objective language cues.

\textit{ModernBERT-large} struggles with subjective classification, achieving 0.41 macro-F1 on train and 0.41 on test, despite showing high objective F1 on test (0.84). After being fine-tuned for 2 epochs with $\alpha$ = 2e-5, it completely failed to capture subjective expressions during validation and test, suggesting overfit and limited adaptability to subjective nuances in the training data.

Among the transfer-learning models, \textit{Sentiment-Analysis-BERT} showed strong performance, attaining 0.87 macro-F1 on train, 0.64 on validation, and 0.67 on test with 4 epochs and $\alpha$ 2e-5. Although validation subjective F1 (0.53) was notably lower than objective F1 (0.71), it generalized better on test with balanced F1 scores (0.58 and 0.77).

\textit{Emotion-English-DistilRoBERTa-base} achieved the best performance among all models with 0.90 macro-F1 on train and 0.77 macro-F1 on validation. It also showed strong and balanced validation scores for both objective (0.78) and subjective (0.75) classification. After being fine-tuned for 6 epochs with $\alpha$ 2e-4, it sustained decent test performance (0.68 macro-F1), although subjective F1 dropped to 0.57. This model was used for submission and placed us $16^{th}$ out of 24.

\textit{Emotion-English-RoBERTa-large} showed similar performance to the distilled model before: 0.67 macro-F1 on test after 7 epochs with $\alpha$ 2e-5. It had similar validation results (0.77 macro-F1), and a very high train performance (macro-F1: 0.97).

\begin{table}[h]
\caption{Results After Fine-Tuning with Augmented Train Data}
\label{tab:results-aug}
\begin{tabular}{l|l|c|c|c|c}
\multicolumn{1}{c}{\multirow{2}{*}{\textbf{Model}}} & {\multirow{2}{*}{\textbf{Dataset}}} &  \textbf{Train} & \multicolumn{3}{c}{\textbf{Validation F1}}\\ 
\multicolumn{1}{c}{} & {} & \textbf{Macro F1} & \textbf{Obj.} & \textbf{Subj.} & \textbf{Macro} \\ \toprule
\multirow{4}{*}{\makecell[l]{Emotion-English-\\DistilRoBERTa-base}} 
& Balanced-2 & 0.95 & 0.62 & 0.68 & 0.68 \\
& Balanced-6 & 0.98 & 0.57 & 0.70 & 0.64 \\
& Corrected Balanced-2 & 0.99 & 0.71 & 0.67 & 0.74 \\
& Corrected Balanced-6 & 0.99 & 0.71 & 0.70 & 0.71 \\ \midrule
\multirow{4}{*}{\makecell[l]{Sentiment-Analysis-BERT}}
& Balanced-2 & 0.91 & 0.64 & 0.65 & 0.67 \\
& Balanced-6 & 0.99 & 0.54 & 0.68 & 0.62 \\
& Corrected Balanced-2 & 0.99 & 0.68 & 0.36& 0.75 \\
& Corrected Balanced-6 & 0.99 & 0.67 & 0.73 & 0.74
\end{tabular}
\end{table}

Table~\ref{tab:results-aug} reports validation results from fine-tuning \textit{Emotion-English-DistilRoBERTa-base} and \textit{Sentiment-Analysis-BERT} on four augmented versions of the training data. Both models achieve high training macro-F1 scores (above 0.90), indicating strong fit to the training data across all configurations.

For \textit{Emotion-English-DistilRoBERTa-base}, the highest validation macro-F1 was achieved with the Corrected Balanced-2 dataset (0.74), where objective and subjective F1 scores were 0.71 and 0.67, respectively. Corrected Balanced-6 performed slightly worse, but remained strong (0.71 macro-F1). The Balanced-2 dataset without editing resulted in a lower macro-F1 (0.68), while Balanced-6 showed the weakest performance overall (0.64), with an especially low objective F1 (0.57).

\textit{Sentiment-Analysis-BERT} showed greater variability. Corrected Balanced-6 yielded the best overall macro-F1 (0.74) with balanced class performance. Corrected Balanced-2 produced a slightly higher macro-F1 (0.75), but this was driven by a strong objective F1 (0.68) and a low subjective F1 (0.36), indicating an imbalance in class predictions. The Balanced-2 dataset without editing led to more balanced class scores (0.64 and 0.65) and a moderate macro-F1 of 0.672. Balanced-6 performed the worst (0.63 macro-F1), with notably lower objective performance (0.54).

In general, editing improved consistency in subjective classification. \textit{Emotion-English-DistilRoBERTa-base} performed reliably across both dataset sizes when editing was applied. In contrast, \textit{Sentiment-Analysis-BERT} was more sensitive to augmentation type and prone to overfitting, particularly favoring objective predictions when trained on the Corrected Balanced-2 dataset.

\section{Discussion}
\label{sec:disc}

Our results highlight the benefits and limitations of transfer-learning of already specified models and data augmentation for subjectivity detection in news text. Among models trained only on the original data, already fine-tuned encoders on related tasks, like \textit{Sentiment-Analysis-BERT} and \textit{Emotion-English-DistilRoBERTa-base}, outperformed general pre-trained models in macro-F1 on validation and test set. In addition, they achieve a better performance over the general-purpose encoders on the test set after fine-tuning. This suggests that pretraining on sentiment and emotion-related corpora improves sensitivity to subjective linguistic cues and confirms \textbf{Hypothesis H1}.

Gains from augmentation were not uniform. While adding more paraphrases (Balanced-6) increased training scores, it did not always translate to better validation performance. In some cases, especially without correction, augmentation degraded performance due to inconsistencies between the generated sentence and its intended label. Clearly, the model performance suffered due to overfit on  noise, which confirms past findings that LLM-generated data can introduce noise without sufficient validation. Therefore, \textbf{Hypothesis H2} needs to be rejected as fine-tuning with augmented train data did not improve performance.

Applying a correction pipeline improved validation macro-F1 for both models. \textit{Emotion-English-DistilRoBERTa-base} showed consistent gains across datasets, with Corrected Balanced-2 and Corrected Balanced-6 outperforming their uncorrected counterparts. \textit{Sentiment-Analysis-BERT} exhibited more volatile behavior. Its macro-F1 improved in Corrected Balanced-6 with balanced class performance, but Corrected Balanced-2 showed a misleading macro-F1 gain driven by high objective performance, while subjective performance dropped substantially. Among all configurations, Corrected Balanced-6 yielded the most stable performance across classes for both models. This suggests that correcting label and style alignment in synthetic data improves generalization and robustness, confirming \textbf{ Hypothesis H3}.

Importantly, these improvements were not just a result of adding more data, but of the generative model self-refining the augmented data to ensure semantic and stylistic alignment. The second-stage editing process using GPT-4o, which rewrote misaligned samples while preserving subject matter and label intent, played a key role in reducing label noise and improving model reliability. This distinction between simple augmentation and corrected augmentation was critical to achieving consistent gains.

Although the models fine-tuned on Corrected Balanced-2 and Corrected Balanced-6 performed best overall, we were not able to submit them to the shared task because results were finalized after the submission deadline. 
We submitted results with \textit{Emotion-English-DistilRoBERTa-base} fine-tuned on the original train dataset which resulted in a 0.68 test macro-F1 and placed us on rank 16 out of 24 in the monolingual-english competition. Therefore, our submission is significantly better than the organizers baseline (test macro-F1: 0.54) but leaves room for improvement due to the top competitor achieving a 0.81 macro-F1 on the test set.
Overall, these findings show that combining transfer-learning with encoders with carefully curated synthetic examples can improve performance in low-resource tasks like subjectivity detection, as long as the synthetic data is reliable and label-consistent.

\section{Future Work}
\label{sec:future}
The research can be extended by applying our approach to multilingual settings, leveraging languages such as Arabic, Bulgarian, German, Italian, Spanish, and French included in Task 1. Our approach can further be improved by incorporating more data, e.g. the dev dataset, into the model fine-tuning for submission. More labeled data can improve the quality of the fine-tuned classifier. Also, the contrastive learning approach can be enhanced by incorporating more specialized loss-functions for the model. Furthermore, we have observed models with varying qualities. For example, \textit{Emotion-english-RoBERTa-large} has demonstrated a greater capability of correctly identifying objective sentences while \textit{Emotion-english-DistilRoBERTa-base} has superior performance for subjective sentences. An ensembling approach of these models could further enhance the classification performance and is an open research avenue. Our contribution to subjectivity detection may be integrated into the broader fact checking framework via misinformation analysis, bias assessment, and claim verification workflows .

\section{Conclusions}
\label{sec:conclusion}
In this paper, we explored subjectivity detection in news text through the lens of transfer-learning and data augmentation. Our findings highlight three key insights: First, domain-specific encoder models already fine-tuned on sentiment or emotion datasets consistently outperformed general pre-trained encoders, supporting our hypothesis that transfer learning can enhance sensitivity to subjective language. Second, while naive data augmentation introduces inconsistencies that occasionally degraded performance, a "self-corrective" pipeline using GPT-4o significantly improved the quality and label alignment of synthetic examples. This allowed models to generalize better and increased macro-F1 scores, particularly for the harder-to-detect subjective class. Third, although our final submission was constrained to uncorrected data due to time limits, our post-submission results demonstrate the value of integrating high-quality synthetic training data. Overall, our approach underscores the importance of not only augmenting data, but shows the ability of a generative model to check itself for semantic and stylistic fidelity. Combining specialized encoders with refined augmentation holds promise for improving low-resource NLP tasks like subjectivity detection.

\section*{Acknowledgements}

We thank the DS@GT CLEF team for providing valuable comments and suggestions.
This research was supported in part through research cyberinfrastructure resources and services provided by the Partnership for an Advanced Computing Environment (PACE) at the Georgia Institute of Technology, Atlanta, Georgia, USA.

\section*{Declaration on Generative AI}
During the preparation of this work, the authors used OpenAI-GPT-4o in order to: Grammar and spelling check. After using this tool, the authors reviewed and edited the content as needed and take full responsibility for the publication’s content.

\bibliography{sample-ceur}

\appendix

\end{document}